\documentclass{article}
\usepackage{preamble}

\title{ Collection Space Navigator: An Interactive Visualization Interface for Multidimensional Datasets
\vspace{-0.1cm}
}

\author{
Tillmann Ohm*‡\textsuperscript{1,}\textsuperscript{2} 
Mar Canet Solà*‡\textsuperscript{1,}\textsuperscript{3} \\
Andres Karjus\textsuperscript{1,}\textsuperscript{4} 
Maximilian Schich\textsuperscript{1,}\textsuperscript{3} \\
\small \textsuperscript{1}ERA Chair for Cultural Data Analytics, Tallinn University  \hspace{1mm} \\
\small \textsuperscript{2}School of Digital Technologies, Tallinn University  \hspace{1mm} \\
\small \textsuperscript{3}Baltic Film, Media and Arts School, Tallinn University  \hspace{1mm} \\
\small \textsuperscript{4}School of Humanities, Tallinn University  \hspace{1mm} \\
\small ‡Corresponding authors: tillmann.ohm@tlu.ee, mar.canet@tlu.ee \\
\small *equal contribution as first authors 
}

\date{\small May 10, 2023 \vspace{-0.4cm}} %
\begin{document}
\maketitle

\begin{abstract}
   We introduce the Collection Space Navigator (CSN), a browser-based visualization tool to explore, research, and curate large collections of visual digital artifacts that are associated with multidimensional data, such as vector embeddings or tables of metadata. Media objects such as images are often encoded as numerical vectors, for e.g. based on metadata or using machine learning to embed image information. Yet, while such procedures are widespread for a range of applications, it remains a challenge to explore, analyze, and understand the resulting multidimensional spaces in a more comprehensive manner. Dimensionality reduction techniques such as t-SNE or UMAP often serve to project high-dimensional data into low dimensional visualizations, yet require interpretation themselves as the remaining dimensions are typically abstract. Here, the Collection Space Navigator provides a customizable interface that combines two-dimensional projections with a set of configurable multidimensional filters. As a result, the user is able to view and investigate collections, by zooming and scaling, by transforming between projections, by filtering dimensions via range sliders, and advanced text filters. Insights that are gained during the interaction can be fed back into the original data via \textit{ad hoc} exports of filtered metadata and projections. This paper comes with a functional showcase demo using a large digitized collection of classical Western art. The Collection Space Navigator is open source. Users can reconfigure the interface to fit their own data and research needs, including projections and filter controls. The CSN is ready to serve a broad community.
\end{abstract}

\section{Introduction}

Large collections of digital artifacts with multidimensional associated metadata can be studied using browsable interactive visualizations that reflect or at least resonate with the intrinsic shape of their data \parencite{manovich2012media}. Mapping the multidimensional topology of a collection into a multidimensional space can help to better understand the overall structure of a dataset and can uncover patterns hinting at underlying trends and dynamics. For example, a researcher or curator may visually explore the space, as constituted by some measure of artifact similarity, looking at different groups of similar objects to identify regions of interest for further quantitative and qualitative investigation.

Multidimensional feature vectors further describe artifact properties in a heterogeneity of ways. This can include both categorical and numerical information. Numerical properties can be derived directly from metadata, such as in the case of artifact creation dates, or constructed through various feature extraction techniques. Neural network methods, for example, can encode measures of complex text semantics \parencite{devlin2018bert}, of visual image properties \parencite{krizhevsky2017imagenet,simonyan2014very,he2016deep,kolesnikov2020big}, of joint image-text pair embeddings \parencite{sohl2015deep,radford2021learning,rombach2022high}, or of spectral audio features \parencite{ren2018deep}. Hand-crafted feature engineering approaches \parencite{zhang2020inkthetics} and more straightforward algorithmic approaches such as compression ensembles \parencite{karjus2022compression} further promise to offer more interpretable vector representations.

Dimensionality reduction techniques can be used to reduce high-dimensional data to a more manageable number of dimensions by remapping or projecting the multidimensional topology into a lower-dimensional coordinate space \parencite{pearson1901liii,van2008visualizing,mcinnes2018umap,amid2019trimap,tang2016visualizing}. For visual interpretation of multidimensional embedding spaces, such projection methods are used to present the data in two or three dimensions, which essentially can function as a more or less distorted reference topography of the original multidimensional topology. The challenge for dimensionality reduction techniques is to preserve complex relationships, while removing information. For example, objects close to each other in high-dimensional topological space should ideally also be close to each other in the low-dimensional topographic projection space. 

Multiple projection views can help to better understand multidimensional data \parencite{amid2019trimap}. Two-dimensional static projections are visually comprehensible, but can only provide a limited view into multidimensional data. Alternatively, to gain intuition over a high-dimensional vector space, it can be helpful to interpret and compare many different projections.
Interactive components in graphical user interfaces (GUI) can further help to gain intuition over the complex interactions between many dimensions. Graphical user interface elements such as range sliders are particularly useful to navigate through multiple features and dimensions, and to query and filter the data \parencite{williamson1992dynamic,ahlberg1994visual}.

\section{Related Work}

Most closely related to the Collection Space Navigator are multiple interactive visualization experiments and prototypes which have emerged in recent years, and which mediate high-dimensional embeddings through low-dimensional projections. They either try to provide an intuitive understanding of multidimensionality and embedding methods by visualizing datasets commonly used for Machine Learning tasks \parencite{smilkov2016embedding, ClouderaFastForwardLabs2019, Custer2020}, or offer explorative interfaces to similarity spaces of cultural collections \parencite{diagne2018t,duhaime2017,oygard2018Visualizing,glinka2017past}. These interactive visualization projects aim to provide overview and deeper insight into their collections, tailored to specific datasets. The VIKUS Viewer \parencite{pietsch2017vikus,pietsch2020Visuelle} offers a more general framework for exploring cultural collections. It allows not only to view a collection as a similarity map of its image embeddings, but also to dynamically filter metadata such as time and categories. The Selfiexploratory of the Selfiecity project \parencite{manovich2015} similarly uses a number of range sliders to filter a multidimensional dataset of images.

The CSN user interface aligns with classic conventions of cultural cartography and modern examples of interactive data visualization and scholarly figure design. Following long-established conventions, the CSN combines a cartesian projection with an auxiliary index and call-out details \parencite{nolli1748nuova}.

Following a modern paradigm of interactive figure design \parencite{bertin2010semiology,victor2011scientific}, the CSN further allows for a deeper functional user experience (UX) and eventually understanding of multidimensional data. The navigation paradigm of the CSN range sliders, which function as Dimension Filters, further resonates, in particular, with the recent state-of-the-art of understanding mathematical multidimensionality via interactive animation \parencite{sanderson2017dimensional}. The CSN combines these foundations with the paradigm of a scatter plot of images as made popular in Cultural Analytics since 2012 \parencite{manovich2012compare}.
The CSN is also functionally similar to network visualization applications, such as Cytoscape \parencite{shannon2003cytoscape} or Gephi \parencite{bastian2009gephi}, which focus on depicting another (yet related) form of multidimensionality in node-link diagrams of complex networks \parencite{bohm2022attraction}.

Even more broadly, the authors of the TensorFlow Embedding Projector \parencite{smilkov2016embedding} suggest to include multipanel projections, i.e. more than one simultaneous projection panel. This would also make sense as a possible extension of the CSN and would be in line with the prevalence of ``multi-chart" figure panels in multidisciplinary science journals \parencite{lee2017viziometrics, berghaus1845physikalischer}.

\section{The Collection Space Navigator}

\begin{figure*}[h]
 	\noindent
 	\includegraphics[width=\columnwidth]{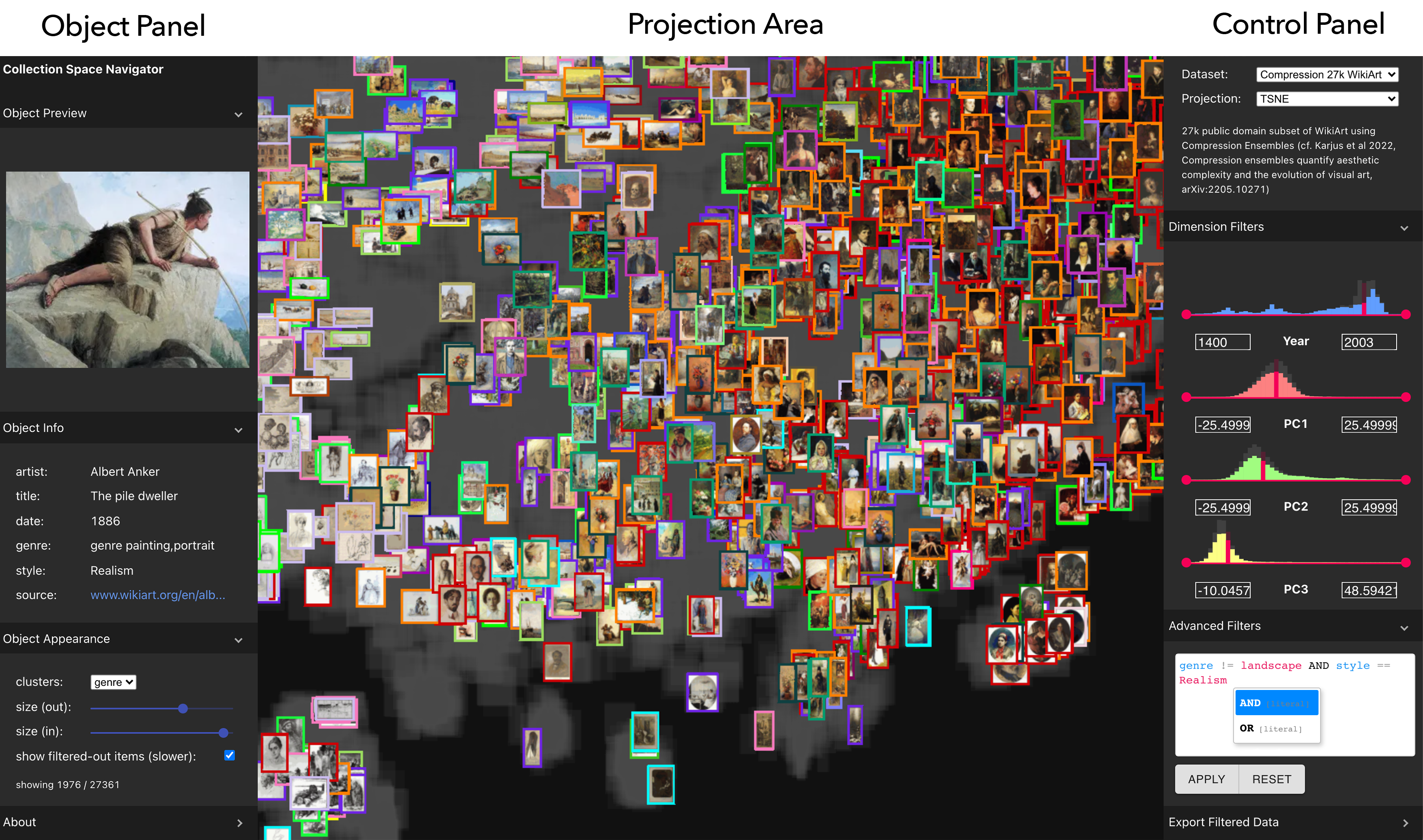}
\caption{
The Collection Space Navigator (CSN). The central \emph{Projection Area} displays a x-y scatter plot of images based on the selected projection (e.g. UMAP, t-SNE), with filtered images greyed out, and mouse-over highlight. The \emph{Object Panel} (left) shows a larger \emph{Object Preview} of the highlighted image, together with \emph{Object Info} based on selected metadata; \emph{Object Appearance} visualizes clusters (optional), sets the projection thumbnail size (zoomed-out and zoomed-in). The \emph{Control Panel} (right) allows for selection of \emph{Data and Projections}; custom interactive \emph{Dimension Filters} and \emph{Advanced Filters} facilitate dataset exploration, analysis, and understanding (see text); the filtered object metadata and current projection view can be downloaded via \emph{Export Filtered Data}.
}
\label{fig_mainfig}
\end{figure*}

\subsection{Motivation}

We developed the Collection Space Navigator as a flexible browser-based research tool applicable across various use cases and research domains: 
\begin{enumerate}
 \item{Researching large collections of digital objects (e.g. images, videos, audio, text, 3D models) with the ability to identify patterns and similar groups based on metadata and vector embeddings;}
 \item{Understanding multidimensionality and projection methods by comparing different embedding spaces and dimensionality reduction techniques through intuitive navigation;}
 \item{Presenting entire media collections online and communicating research findings with diverse audiences.}
\end{enumerate}
While prototypes and use cases exist for each of these aspects as discussed above, we are not aware of a tool that meets all of these requirements. Our contribution therefore lies in the combination and extension of existing interfaces to work towards a more universal and open modular research and curation system.

\begin{figure*}[h]
\noindent
\includegraphics[width=\columnwidth]{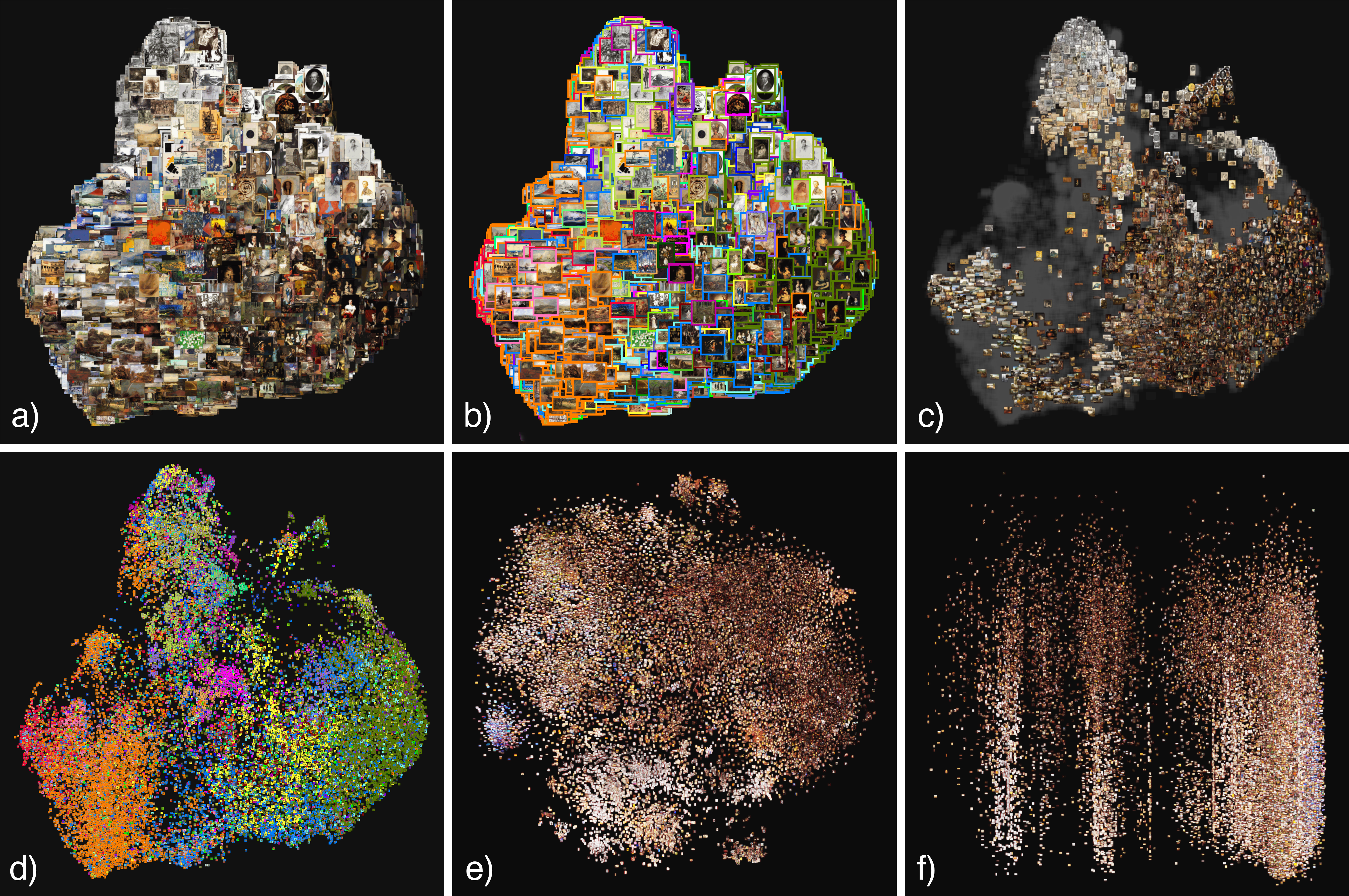}
\caption{
Examples of various 2D projections and visualization features in the CSN tool. a) UMAP projection with large thumbnails, providing a comprehensive view of the image content; b) UMAP projection with medium size thumbnails and cluster colors of categorical data selected in \emph{Object Appearance}; c) UMAP projection with medium size thumbnails and filtered-out objects in grey; d) UMAP projection with small thumbnails and cluster colors, providing a more compact representation; e) t-SNE projection, showing an alternative dimensionality reduction technique; f) Simple x/y plot, here showing PC1 over time for temporal analysis. The flexibility of the CSN tool provides the ability to effectively explore and compare data across different visualization methods via displaying multiple different 2D projections, flexible based on data import configuration, selectable in \emph{Data \& Projections}, combined with thumbnail sizes and cluster highlights as adjustable via the \emph{Object View Settings}.
}
\label{fig_projectors}
\end{figure*}

\subsection{Design principles}

To make complex interactions in multifaceted datasets comprehensible while meeting the diverse needs of researchers, we formulated three design principles for the tool: 
\begin{enumerate}
\item{Providing an open modular system that adapts to different research needs, domains and datasets while preventing information overload;}
\item{Providing a complete overview of the collection while encouraging immersive exploration of the objects;}
\item{Providing a multitude of interaction mechanisms and modalities to foster intuition, such as zooming, panning, hovering, and sliding through feature dimensions.}
\end{enumerate}

\subsection{Components}

\subsubsection{Projection Area}

The central part of the interface is the \emph{Projection Area} (Figure  \ref{fig_mainfig} center). It maps the given input collection in its entirety as miniature images in an interactive 2D scatter plot, with coordinates defined by the chosen projection method. Basic navigation operations such as zooming or ``drag and move" allow free exploration of the projection space. While the user sees only 2 dimensions in the central projection area, the CSN technically includes a third axis for depth. Moving along the depth axis (by zooming) effectively reveals overlapping objects. The appearance of the thumbnails can be adjusted in the \emph{Object Panel}.

\subsubsection{Object Panel}

The \emph{Object Panel} (Figure \ref{fig_mainfig} left) has three collapsible sub-menus: \emph{Object Preview}, \emph{Object Info} and \emph{Object Appearance}. 
The \emph{Object Preview} section displays a larger version of the miniature thumbnail currently hovered on in the \emph{Projection Area}. 

By default it simply shows a larger version of the same thumbnail, but it can be set to display a higher resolution version of the hovered image, stored either locally or remotely.

The \emph{Object Info} section provides detailed information on the currently selected object. This aspect of the CSN is highly flexible, as the metadata fields that provide this information can be easily defined in the separate configuration file. Minimally it can display just the file name, but it can equally well include extensive metadata --- for example in the case of art collections, the author, production year, location, genre, style, and other details.

The \emph{Object Appearance} section contains options to control the visual appearance of the objects in the \emph{Projection Area}. A predefined group (from categorical metadata) can be selected from a drop-down list to show clusters. Objects of the same category are depicted with the same color border around their thumbnails. The size and scale of the miniature images are adjustable with convenient sliders. Size determines how large the thumbnails should be when fully zoomed out, while Scale affects the size when zoomed in.

\subsubsection{Control Panel}
 
The \emph{Control Panel} (Figure \ref{fig_mainfig} right) has four collapsible sub-menus: \emph{Data \& Projections}, \emph{Dimension Filters}, \emph{Advanced Filters}, and \emph{Export Filtered Data}. 
The \emph{Data \& Projections} section contains a \emph{Dataset} drop-down list of selectable datasets. For very large collections, we recommend providing a smaller subset by default and offering the entire set on demand (such subsets can be conveniently produced using the CSN configuration Python notebook). The section also contains a \emph{Projection} drop-down list of selectable projections and mappings. Switching between different projections, e.g. different embedding or reduction methods, smoothly rearranges the positions of the objects in the \emph{Projection Area}. These intuitive animations can provide new insights into the intermediate state between two projections and expose their differences.

\begin{figure}[h]
 	\noindent
 	\includegraphics[width=\columnwidth]{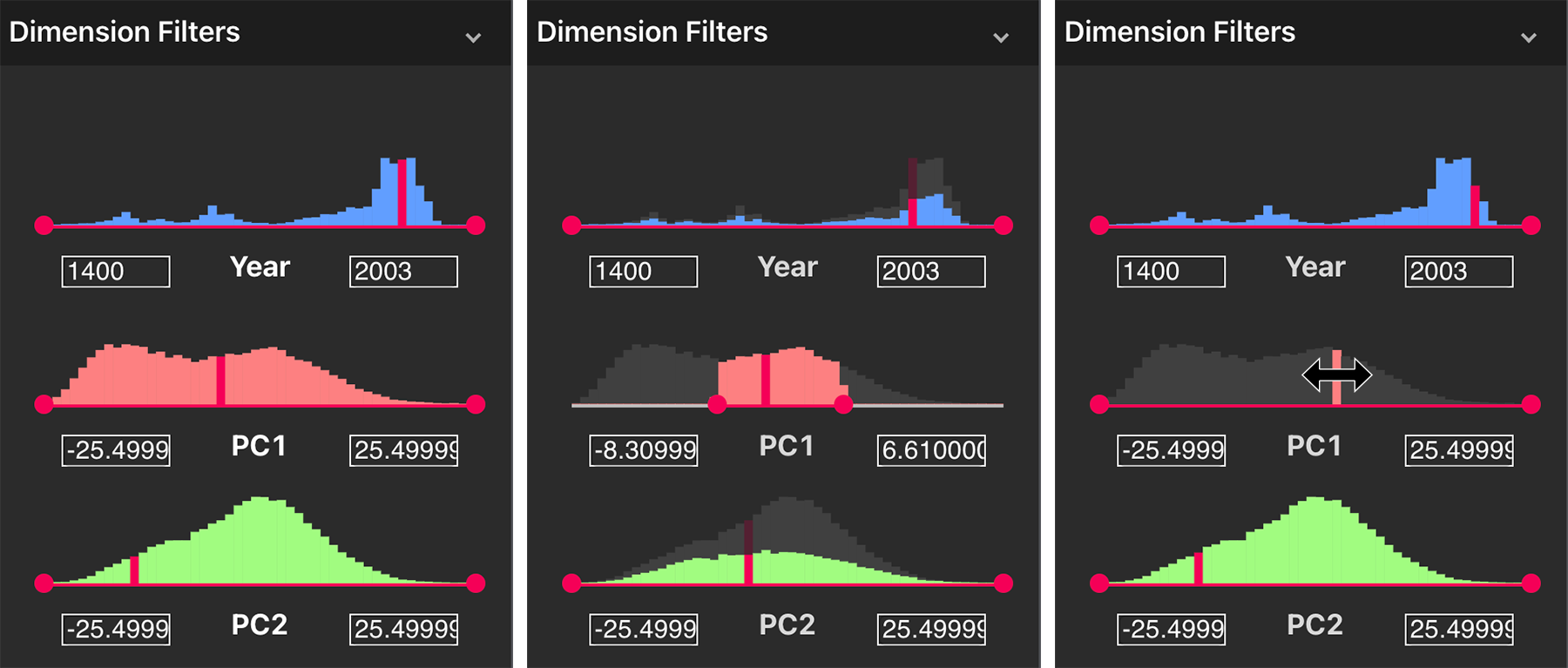}
 	\caption{
  Interactive Dimension Filters. Left: Unfiltered Dimension Filters, consisting of range sliders with interactive histograms above them, showing the distribution of all objects along the slider’s dimension with the bin of the currently selected object highlighted in red; Center: Reducing the range of one slider affects the distribution of all dimensions, reflected by the histograms; Right: \emph{Bin Mode} functionality is activated by clicking on a histogram, allowing the user to activate one bin at a time, with the Projection Area displaying the corresponding objects within the active bin. These interactive Dimension Filter features allow in-depth exploration and visualization of multi-dimensional data distributions allowing users to gain a deeper understanding of the relationships between data points.
  }
  \label{fig_filters}
\end{figure}

The \emph{Dimension Filters} are optional interactive elements that function to filter the objects in the \emph{Projection Panel} (Figure \ref{fig_filters}). They control the range of the assigned variables, which could be dimensions of the embedding, metadata such as date or year of creation of an artwork, or inferred properties of the image such as colorfulness or contrast. 
Histograms above the range sliders provide additional statistical information and feedback on how changes affect the distribution of the mapping. They are constantly updated to reflect the distribution of the entire dataset as well as the distribution of filtered and unfiltered objects. Clicking on a histogram activates the \emph{Bin Mode}: moving the cursor over the bars of the histogram temporarily displays only objects within the narrow range of the bar. A second click terminates this function and sets the filters back to the previous state. Additionally, hovering over the thumbnails in the \emph{Projection Area} highlights the corresponding vertical bar in each histogram.

The \emph{Advanced Filters} section is a text field to construct and apply search and filter queries. By default, it handles basic query operators such as \texttt{AND}, \texttt{OR}, equals (\texttt{==}), does not equal (\texttt{!=}), as well as custom operators. Nested and complex search queries are also supported using round brackets. In the setup, using the configuration Python notebook, each metadata field can be defined as a Free Text Entry (enabling queries) or as a Categorical Selection (generating drop-down lists, i.e. GUI elements that allow simple search and selection).

The \emph{Export Filtered Data} section in the \emph{Control Panel} allows downloading the metadata of the currently filtered objects as a CSV file, and the current projection view as PNG file.

\section{Applications}

An important aspect of the CSN is the flexibility to combine different data types and different methods for feature extraction and dimensionality reduction. 
To initially test the CSN towards many potential applications in other disciplines, we utilized the application in several use cases within our multidisciplinary research lab. All these cases vary in terms of research questions, approaches, and datasets. Here we briefly describe three examples which (i) analyze embedding spaces, (ii) study multidimensional metadata, and (iii) explore and curate large sets of generated images respectively. Mixed applications are of course possible. A growing diversity of application demos will be added to the project website\footnote{https://collection-space-navigator.github.io/}.

\subsection{Analyzing aesthetic complexity in art collections}

Karjus et al. (2022) propose quantifying visual aesthetic complexity using ``compression ensembles", an information theory driven method based on transforming images using visual filters and repeatedly compressing them. This approach produces embeddings analogous to those from deep learning based image embeddings, but with explainable and arguably more interpretable results in the aesthetic domain.
This paper uses the same dataset as our examples in figures 1 to 3 and our associated initial demo application, i.e. a corpus of digitized images of visual art such as paintings and drawings, a subset of 74,028 works of the art500k project corpus \parencite{mao2017deepart}, in turn primarily sourced from Wikiart, covering mainly Western art of the last 400 years. 

Here and in similar projects focusing on visual art, the CSN can serve both as a research and a presentation tool in several capacities. The visual and interactive overviews of the embedded artworks allow for a closer examination of art historical trends over time and in relation to their aesthetic complexity. Using multiple representations and filters can further help to build an intuitive understanding of the given embedding method. In the case of compression ensembles, the variables or dimensions are interpretable \parencite{karjus2022compression}, and can be decorrelated and grouped, for example using Principal Component Analysis. A concrete usage of the CSN in this circumstance is for example to display the dataset in the projection area using two principal components or specific compression dimensions as x and y coordinates, while filtering another compression dimension with the range sliders to visualize the respective interaction. 

\begin{figure*}[h]
 	\noindent
 	\includegraphics[width=\columnwidth]{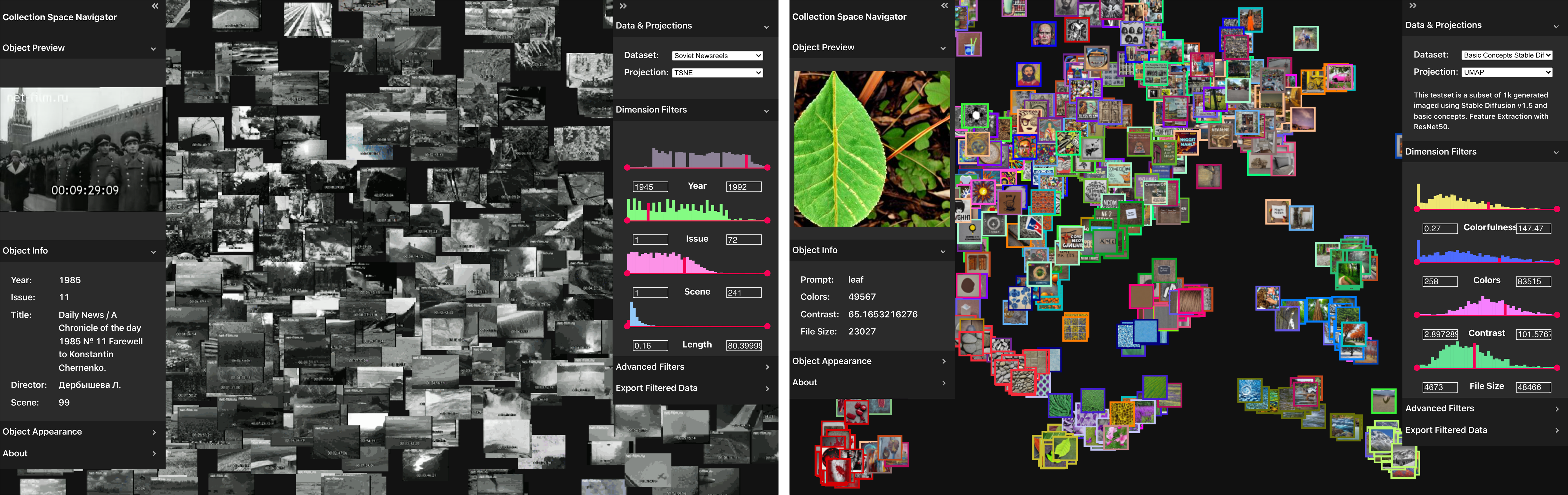}
 	\caption{
Two distinct use cases of the Collection Space Navigator. Left: Historical newsreels with rich metadata; Right: Investigating text-to-image generation through Stable Diffusion with linguistic basic concepts as prompts. By using domain specific metadata filters and projections, the CSN tool allows for deeper exploration providing insight into visual and thematic patterns. 
  }
  \label{fig_newsreel}
\end{figure*}

\subsection{Studying represented worldviews in historical newsreels}

Oiva et al. (forthcoming) propose a multidisciplinary framework for the study of historical newsreels, which can be extended to news broadcasts and audiovisual collections more broadly. The initially analyzed dataset consists of 1747 newsreels produced weekly over 50 years in the Soviet Union.

Within the project, the digitized video clips, about 8-10 minutes each, are split into individual frames, i.e. static images, which form the unit of analysis either in their full granularity or coarse-grained, for example as a single frame per shot. The material consists of predominantly black and white footage of variable quality, transitioning to color towards the end of the period. Rich metadata accompanies the visual information. Taking all this into account, the authors aim to analyze and understand the material, including in particular the variation of intrinsic worldviews.

In this project, the CSN is used both as a presentation tool to communicate to the audience, and a research tool for the authors (Figure \ref{fig_newsreel}), allowing for exploration of the film footage, to identify recurring patterns and shots, and to generate hypotheses based on a combination of the visual representation of movie frames based on image embedding vectors, and accompanying metadata. In this case, indeed, a stronger focus lies on the latter explicit metadata-dimensions, which can be mapped to range-sliders in a similar way as the numerical vectors in example 4.1.

\subsection{Exploration and curation of text-to-image models}

The advent of easily accessible text to image models such as DALL-E, Midjourney, and Stable Diffusion \parencite{rombach2022high}. %
has brought artistic and photo-realistic image generation to the mainstream.
This is very likely to transform creative industries in the near future. Yet the effectively near-infinite content capacity will require understanding of the models to be made use of, and efficient curation to work with already generated images. 

The CSN is well suited for such tasks, allowing for insight into large image sets using various visual similarity and metadata representations and filtering options. For the demo, we generated of images with Stable Diffusion 1.5 using the basic word list of the Automated Similarity Judgement Program (ASJP) \parencite{brown2008automated} as the input text prompts, and embedded them using the ResNet50 pretrained image model. 

This approach, querying simple terms such as colors, body parts and animals, allows for probing the ``defaults" of the text2image models, while generating multiple versions of each prompts yields insight into variation in its capacity. The same approach could be used to compare different models, as well as investigate effects of complexity of the prompts. Naturally, more complex prompts such as those common in artistic text2image creation could be explored as well.

\section{Conclusion}
The Collection Space Navigator (CSN) as introduced in this paper is a powerful and flexible data visualization tool working with large sets of images associated with multidimensional data, including embedding vectors and tabular metadata. The CSN is publicly available on Github, including a functioning demo as well as a How-To configuration and usage guide in form of an interactive Python notebook. As described above, the interface can be conveniently and readily tailored to the needs of a particular research project and dataset. An important aspect of the CSN in this regard is its flexibility to combine different vector and metadata types, a variety of projection methods, and a diversity of representations. This includes the bespoke configuration of different representation and filter controls, resulting in an elegant user interface that allows the user to focus on the material under investigation, without unnecessary distractions in the form of "default" controls.

We have exemplified the CSN here using visual data. Yet, while input data is by default represented by small images or thumbnails in the Projection Area, the input does not need to be vectors derived from images. For example, embeddings or metadata of audio or text could be visualized equally well, represented by suitable thumbnails or labels. We imagine, such "unorthodox" usage could facilitate the exploration and further research on such non-visual spaces of meaning too. By making the code fully open source, we encourage further development in this and other directions, for example adding playback functionality to images associated with audio or video files. 

In sum, the Collection Space Navigator (CSN) is a new research tool that facilitates the work of artists, curators, scholars, and scientists, who study large collections of digital visual artifacts associated with metadata and/or vector embeddings. We release the CSN as open source under MIT License, to enable broad reuse and further development without restriction. We hope it will invigorate a diverse ecology of multidisciplinary research and curatorial practice, and help to deepen our collective understanding of multidimensional meaning spaces.

\section*{Code availability and demo} 
The CSN is released as open-source (MIT licence), with code and documentation available on GitHub at \url{https://github.com/Collection-Space-Navigator/CSN} 
\newline
A live demo, using the the same example data as in figures 1 to 4 in this paper, is available at \url{https://collection-space-navigator.github.io/CSN}

\section*{Author contributions, acknowledgments and funding} 

T.O. and M.C. designed, co-authored, and developed the Collection Space Navigator. T.O., M.C., A.K. and M.S. contributed to the research design and co-wrote the manuscript. T.O., M.C. and A.K. collected data. T.O. and M.C. contributed equally to this work as first authors. The authors thank Sebastian Ahnert, Mila Oiva, and the entire CUDAN team for useful conversations and input. All authors are supported by the CUDAN ERA Chair project, funded through the European Union's Horizon 2020 research and innovation program (Grant No. 810961)

\begingroup
\setlength{\emergencystretch}{8em}
\printbibliography
\FloatBarrier
\endgroup

\end{document}